\documentclass[10pt, a4paper]{article}
\usepackage{lrec2022} 
\usepackage{multibib}
\newcites{languageresource}{Language Resources}
\usepackage{graphicx}
\usepackage{tabularx}
\usepackage{soul}
\usepackage{titlesec}
\titleformat{\section}{\normalfont\large\bfseries\center}{\thesection.}{1em}{}
\titleformat{\subsection}{\normalfont\SmallTitleFont\bfseries\raggedright}{\thesubsection.}{1em}{}
\titleformat{\subsubsection}{\normalfont\normalsize\bfseries\raggedright}{\thesubsubsection.}{1em}{}
\renewcommand\thesection{\arabic{section}}
\renewcommand\thesubsection{\thesection.\arabic{subsection}}
\renewcommand\thesubsubsection{\thesubsection.\arabic{subsubsection}}

\usepackage{epstopdf}
\usepackage[utf8]{inputenc}

\usepackage{hyperref}
\usepackage{xstring}
\usepackage{enumitem}

\usepackage{color}

\usepackage{comment}
\usepackage[utf8]{inputenc}

\newcommand\blfootnote[1]{%
  \begingroup
  \renewcommand\thefootnote{}\footnote{#1}%
  \addtocounter{footnote}{-1}%
  \endgroup
}

\title{ \vspace*{.5\baselineskip} Hollywood Identity Bias Dataset: A Context Oriented Bias Analysis of Movie Dialogues}
\name{Sandhya Singh$^{\ast}$, Prapti Roy$^{\ast}$, Nihar Sahoo$^{\ast}$,
Niteesh Mallela$^{\ast}$, Himanshu Gupta$^{\ast}$,\\ {\bf \large Pushpak Bhattacharyya$^{\ast}$, 
  Milind Savagaonkar$^{\dagger}$, Nidhi$^{\dagger}$, Roshni Ramnani$^{\dagger}$,} \\ {\bf \large  Anutosh Maitra$^{\dagger}$,      Shubhashis Sengupta$^{\dagger}$}}

\address{$^{\ast}$CFILT, Indian Institute of Technology Bombay, India\\$^{\dagger}$Accenture Labs, India\\
         \{sandhyasingh, nihar, niteesh, pb\}@cse.iitb.ac.in,\\ gupta\_himanshu@iitb.ac.in, praptir1994@gmail.com, nidhidr18@gmail.com\\ \{milind.savagaonkar, roshni.r.ramnani, anutosh.maitra, shubhashis.sengupta\}@accenture.com, \\
         \\
         \\}

\abstract{ 
\textit{\textbf{Warning:} This paper contains content that may be offensive or upsetting however this cannot be avoided owing to
the nature of the work.} \\ \newline 
Movies reflect society and also hold power to transform opinions. Social biases and stereotypes present in movies can cause extensive damage due to their reach. These biases are not always found to be the need of storyline but can creep in as the author's bias. Movie production houses would prefer to ascertain that the bias present in a script is the story's demand. Today, when deep learning models can give human-level accuracy in multiple tasks, having an AI solution to identify the biases present in the script at the writing stage can help them avoid the inconvenience of stalled release, lawsuits, etc. Since AI solutions are data intensive and there exists no domain specific data to address the problem of biases in scripts, we introduce a new dataset of movie scripts that are annotated for identity bias. The dataset contains dialogue turns annotated for (i) bias labels for seven categories, viz., \textit{gender, race/ethnicity, religion, age, occupation, LGBTQ, and other, which contains biases like body shaming, personality bias,} etc. (ii) labels for sensitivity, stereotype, sentiment, emotion, emotion intensity, (iii) all labels annotated with context awareness, (iv)
target groups and reason for bias labels and (v) expert-driven group-validation process for high quality annotations. We also report various baseline performances for bias identification and category detection on our dataset.
 \\ \newline  
 }

\begin{document}

\maketitleabstract

\section{Introduction}
\label{sec:intro}

Movie is an audio-visual form of art dealing with the complex human psyche. It nourishes and nurtures the social ethos and delves deep into the reality of society. Stemming from the social fabric, they have the potential to influence a predominantly large section of public opinion. Social identity has shaped one's life and thus, has emerged as vital for survival. Thus, social biases and stereotypes in movies can lead to significant negative impacts. Identifying bias in natural language processing (NLP) tasks have seen a spurt in recent times. Bias 
brings along the notion of harm to the task in consideration \cite{bishop2006pattern,caliskan} \blfootnote{* equal contribution}. In contemporary times, the presence of biases acts as a major deterrent toward the path of equality. In this work, we define
bias as a prejudice in favor or against a person, group, or community directed towards their social identity. 

Biases in movies emerge not only as the demand of storyline but also, can be infused through screenwriter's own bias. 
The movie production houses prefer to ascertain that bias present in a script is the story demand as the content might lead to controversy and cause inconvenience and losses. To recall, movies like \textit{The Birth of a Nation, The Last Temptation of Christ}\footnote{\url{https://en.wikipedia.org/wiki/List\_of\_banned\_films}} have faced controversies due to their content. Hence, the production houses are cautious about filtering out the harmful utterances at the initial scripting level. The scripts go through multiple iterations for content verification before production. This process is highly human-intensive, involving human judgment and efforts to moderate it.

With Deep Learning (DL) models approaching human-level accuracy in various tasks, an AI-supported solution to identify the biases present in the script at the writing stage is the need of the hour. This can speed up the entire scripting process and ease human effort. Data is prime to DL models, but there is no known available dataset for identifying the biases in the movie script domain.  

Most of the recent works have focused on detection of harmful biases in Hate Speech, offensive texts, and social media posts \cite{kiritchenko2018examining,diaz2018addressing,mathew2020hatexplain}. However, bias detection in movie domain is less explored
apart from few research on gender asymmetry and stereotype \cite{sap-etal-2017-connotation,Kagan2020,Garcia14genderasymmetries,cinderlla}. Some works are also available based on linguistic features to identify gender, race, and age differences in movies and fiction \cite{fast2016shirtless,ramakrishna-etal-2015-quantitative,ramakrishna-etal-2017-linguistic}.
No attention has been given for the detection of biases and stereotypes at the dialogue level for the movie domain. Also, developing a dataset on bias detection for movie dialogues is a time-consuming and complicated. It requires both social and computational expertise. This motivated us to create a dataset to address the problem of social bias identification in movie dialogues.


We introduce a new dataset as Hollywood Identity Bias Dataset (HIBD) consisting of $35$ movie scripts annotated for multiple identity biases. The annotation is done at sentence level with a total of $49117$ sentences, of which $1181$ biased sentences were detected, which amounts to approximately $2.5\%$ bias.
This highlights the challenge involved in annotating the biases in the movie script domain, \textit{viz.,} skewness of the data. 

Besides individual opinions, social biases also emerge from preconceived generic notions of stereotypical perception. During annotation, we found that there are dialogues that are not biased in content but are sensitive in nature. 
 Though researchers have synonymously used bias and stereotypes \citelanguageresource{nadeem-etal-2021-stereoset}, we consider them as separate socio-linguistic phenomena. A detailed discussion is given in \hyperref[sec:data]{section 3} of this paper.



To address all these aspects related to biases, 
we are following a layered approach for annotation.
\textbf{Our contribution} through this work is listed as:
\begin{itemize}[noitemsep] 
    \item To the best of our knowledge, we propose the first context aware, dialogue level bias analysis of Hollywood movies.
    
    \item We have annotated the scripts for Sensitivity, Stereotype and social bias labels as \textit{Gender, Race/Ethnicity, Religion, Age, Occupation, LGBTQ, and Other, that contains biases like body shaming, personality bias}, \textit{etc.}
    \item Each annotated bias is further labeled Implicit or Explicit to convey the nature of bias along with their corresponding target group and the rationale behind it. 
    \item We are annotating sentiment as positive or negative and its associated emotion and intensity based on plutchik's emotion wheel \cite{plutchik2013theories}.
    \item Baseline experiments to identify biases and their category detection is done to benchmark our dataset.
\end{itemize}
The paper is structured as follows: \hyperref[sec:related-work]{Section 2} discusses the related work for bias identification, datasets related to various social biases and why they are not suitable for our problem. \hyperref[sec:data]{Section 3} describes our dataset at length in terms of definition and annotation process. \hyperref[sec:challenges]{Section 4} discusses some of the challenges encountered during annotation and our approach to resolving them with expert intervention. \hyperref[sec:experiment]{Section 5} presents baseline experiments, results and error analysis followed by \hyperref[sec:conclusion]{conclusion and future work} in section 6. The annotated dataset, annotation guidelines and code has been made available\footnotemark. 
\footnotetext{\url{https://github.com/sahoonihar/HIBD_LREC_2022}}

\section{Related Work}
\label{sec:related-work}

\textbf{Bias in NLP:} There has been a lot research on bias detection in various forms of text, starting from word embeddings \cite{caliskan}. \newcite{Bolukbasi} has shown that embeddings like Glove \cite{glove} and Word2vec \cite{word2vec} exhibit human-like gender bias for various occupations. This was extended by \newcite{manzini-etal} to show racial and religious biases. Concurrent works like  \newcite{kurita2019measuring} and \newcite{may-etal} show that contextual embeddings like BERT \cite{devlin-etal-2019-bert} not only encode multiple social biases but also amplify them. The presence of bias in these models and embeddings poses many harmful risks by unfair allocation of resources against the marginalized section of the society such as LGBTQ \cite{blodgett-etal-2020-language}.

Hence, the research community, in its endeavor toward EthicalAI, has seen an upsurge in many supervised datasets, methodologies to detect and mitigate various social biases through different extrinsic tasks like hate speech and toxicity detection \cite{sap-etal-2019-risk,davidson-19,dixon}, coreference resolution \cite{gap}, question answering \cite{qa-bias}, machine translation \cite{gender-bias-mt}, and so on. However, most of the works focus on gender and race biases. Other biases like
religion, LGBTQ, occupation, and body-shaming are still not explored much.

The recently released StereoSet dataset \citelanguageresource{nadeem-etal-2021-stereoset} was crowd-sourced to measure stereotypical gender, profession, race, and religion biases. Another crowd-sourced dataset called CrowS-Pairs \citelanguageresource{nangia2020crows} was published as a challenge set for measuring the degree to which nine types of social bias are present in language models. It has data related to nine bias types: \textit{race, gender, sexual orientation, religion, age, nationality, disability, physical appearance and socioeconomic status}. While these datasets consider multiple dimensions of social biases, a recent work \cite{blodgett-etal-2021-stereotyping} has reported many pitfalls in annotations.

\begin{figure*}
  \centering
  \frame{\includegraphics[width=0.8\linewidth, height=5cm]{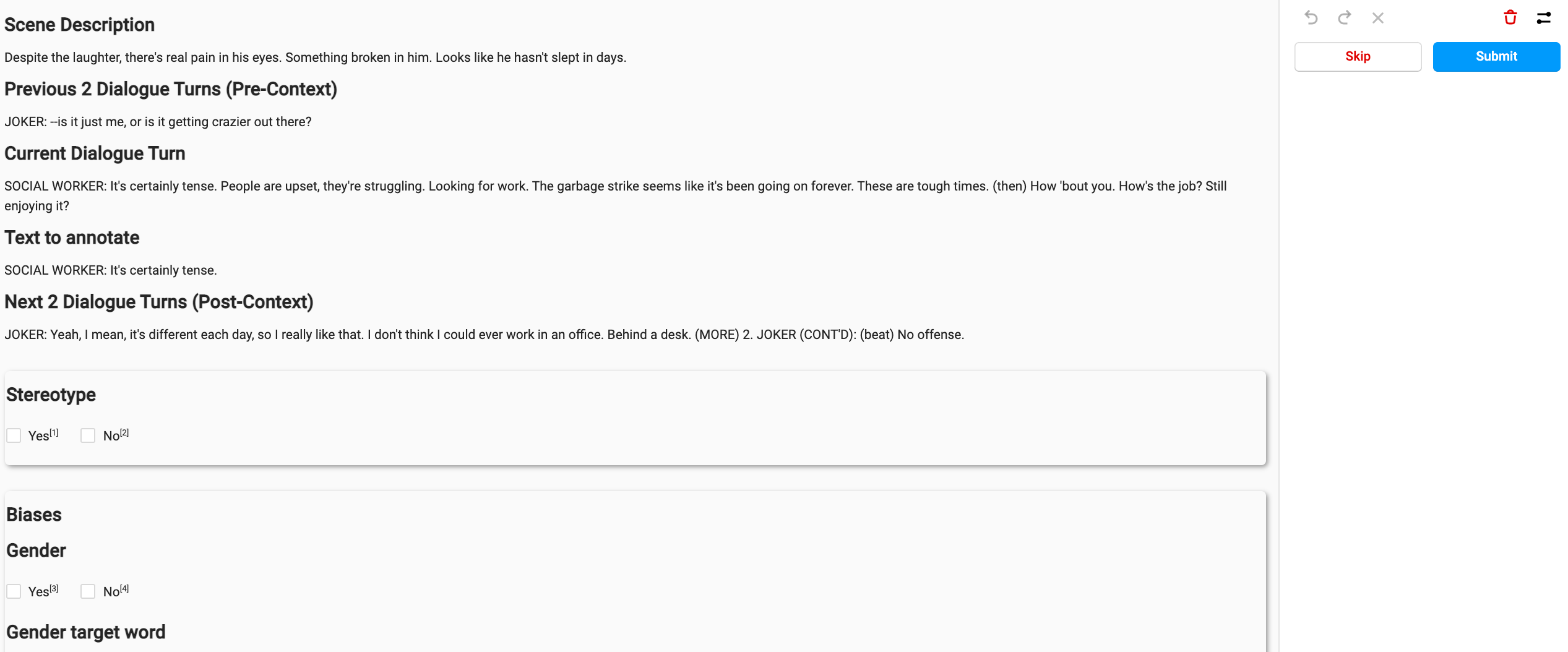}}
   \caption{A Screenshot of the annotation interface showing a Dialogue Turn for annotation }
      \label{fig:interface}
 \end{figure*}

Another work in the direction of defining and evaluating social biases \citelanguageresource{sap2020social} creates a structured way of representing the biases and offensiveness implied in language. The frames combine categorical knowledge about the offensiveness, intent,
and targets of statements, along with free-text reasoning about which groups are targeted and biased implications or stereotypes. They have annotated the dataset for seven bias types: gender/sexuality, race/ethnicity, religion/culture, social/political, disability, body/age, and victims. Bias is a complex phenomenon that requires context information and speaker information to be understood by a human being. However, most previous works have not used any social context to capture these biases. A relatively new work \citelanguageresource{vidgen2021introducing} has annotated English Reddit entries for Identity-directed, Person-directed, Affiliation-directed, CounterSpeech, Non-hateful Slurs, and Neutral categories for abusive text considering the context of the conversation thread. The rationales for annotation are marked in the form of span highlighted in the text.

\textbf{Bias in Movies Scripts:} Though there have been numerous works towards detecting social biases in texts, much focus has not been given to detecting them in the entertainment domain. Different studies \cite{sap-etal-2017-connotation,Kagan2020,Garcia14genderasymmetries,cinderlla} show the gender asymmetry and stereotypes in social media and movies. \newcite{fast2016shirtless}, \newcite{ramakrishna-etal-2015-quantitative} and \newcite{ramakrishna-etal-2017-linguistic} rely on linguistic features to quantify gender, race and age differences in movies and fictions. Unlike other previous works, \newcite{khadilkar2021gender} has made use of Cloze tests and WEAT \cite{caliskan} metrics to measure gender bias along with other subtle biases and compare these biases between two major movie industries of Hollywood and Bollywood. 

But, there has been very little work on identifying biases other than gender in movie scripts. Also, no prior work focuses on the detection of biases and stereotypes at dialogues level for any film industry. We need to make a conscious effort towards automatic detection and classification of biased dialogue turns in movies as their presence may have adverse impact on viewers \cite{DIMNIK2006129,cape}.

This motivated us to develop a data resource that can address the problem of identifying context aware social bias at dialogue level in movie scripts.



\section{Hollywood Identity Bias Dataset}
\label{sec:data}

This dataset is developed primarily to address the problem of identity related social biases and stereotype identification from the entertainment domain. Our objective was to develop a comprehensive labeled dataset for multitasking that can comply with inference, categorization, and classification tasks related to the entertainment domain. 

\subsection{Terminology}
In previous works, biases and stereotypes have often been used interchangeably, thus, lacking clarity. To enable the models to interpret the biased constructs in scripts, a pragmatic formalism was defined to help us differentiate between these related yet different terms. The terms used in our annotation process are defined as:
 
 \begin{itemize}
\item \textbf{Sensitivity:} The \textit{property }of a statement targeted towards an individual or a group belonging to a section that is either overrepresented, underrepresented or vulnerable due to identity such as \textit{race, religion, occupation, sexual identity etc.}  A sensitive utterance can be both implicit and explicit in nature and is loaded with potential tension which might lead to aggression. The notion of sensitivity is deeply rooted in the power position or people power and always bear a negative sentiment. The definition of sensitivity encompasses an array of terminologies such as hate, offensive and abusive text provided an identity is targeted.
For example, \textit{"The Jews are again using holohoax as an escape to spread their agenda."} is a sensitive statement against Jews.


\item \textbf{Stereotype:} It is an \textit{overgeneralized belief} about a particular community. It guides our thought process by prompting us to ignore facts and thus cause biases. When a fact (existential quantifier) is replaced with an overgeneralized statement (universal quantifier), it becomes a stereotype. For example, \textit{"Some white men can not dance"} is a fact statement; but \textit{"White men can not dance"} is a stereotype.

\begin{figure}[h]
  \centering
  \frame{\includegraphics[width=0.85\linewidth, height=1cm]{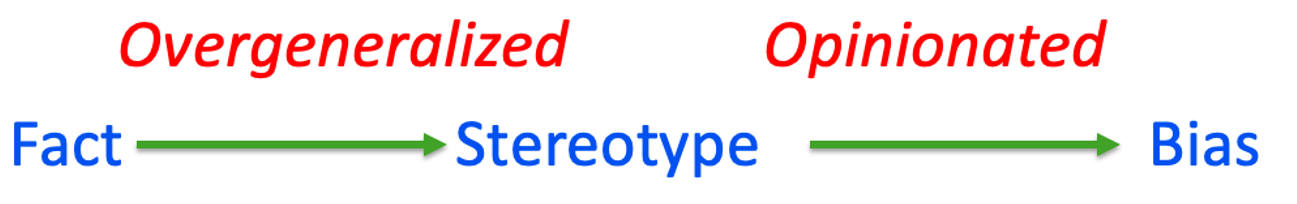}}
   \caption{Relation between fact, stereotype and bias statement. 
   }
      \label{fig:stereotype}
 \end{figure}

\item \textbf{Bias:}
It is a prejudice \textit{towards} or \textit{against} an individual or community. It exists in both conscious and unconscious forms and influences decision making. As given in figure 1, when an overgeneralized idea gets opinionated, it becomes a bias. However, biases do not always originate from stereotypes, and they can be an opinionated statement targeted towards an identity. 
We define bias as a quintuple \textit{$<$S, L, T, C, R$>$} where,
\begin{itemize}
    \item \textit{S} is the communicator (speaker, author)
    \item \textit{L} is the communicatee (audience, reader)
    \item \textit{T} is the target of the bias 
    \item \textit{C} is the identity category of bias 
    \item \textit{R} is the reason for bias

\end{itemize}

For example, in the following dialogue set:

\fbox{\begin{minipage}{15em}
\textit{\textbf{KIRSTY:} She's so damn...English.} \\ \textit{\textbf{STEVE:} Meaning what?}
\end{minipage}}\\

  \textit{Kirsty} is the communicator, \textit{Steve} is the communicatee, Target is \textit{English}, Bias Category is \textit{race} and the reason for bias is that \textit{"British people are believed to be overly controlled community"}

\end{itemize}

\subsection{Structure of a Movie Script}

\begin{figure}[!h]
  \centering
  \frame{\includegraphics[width=\linewidth, height=5cm]{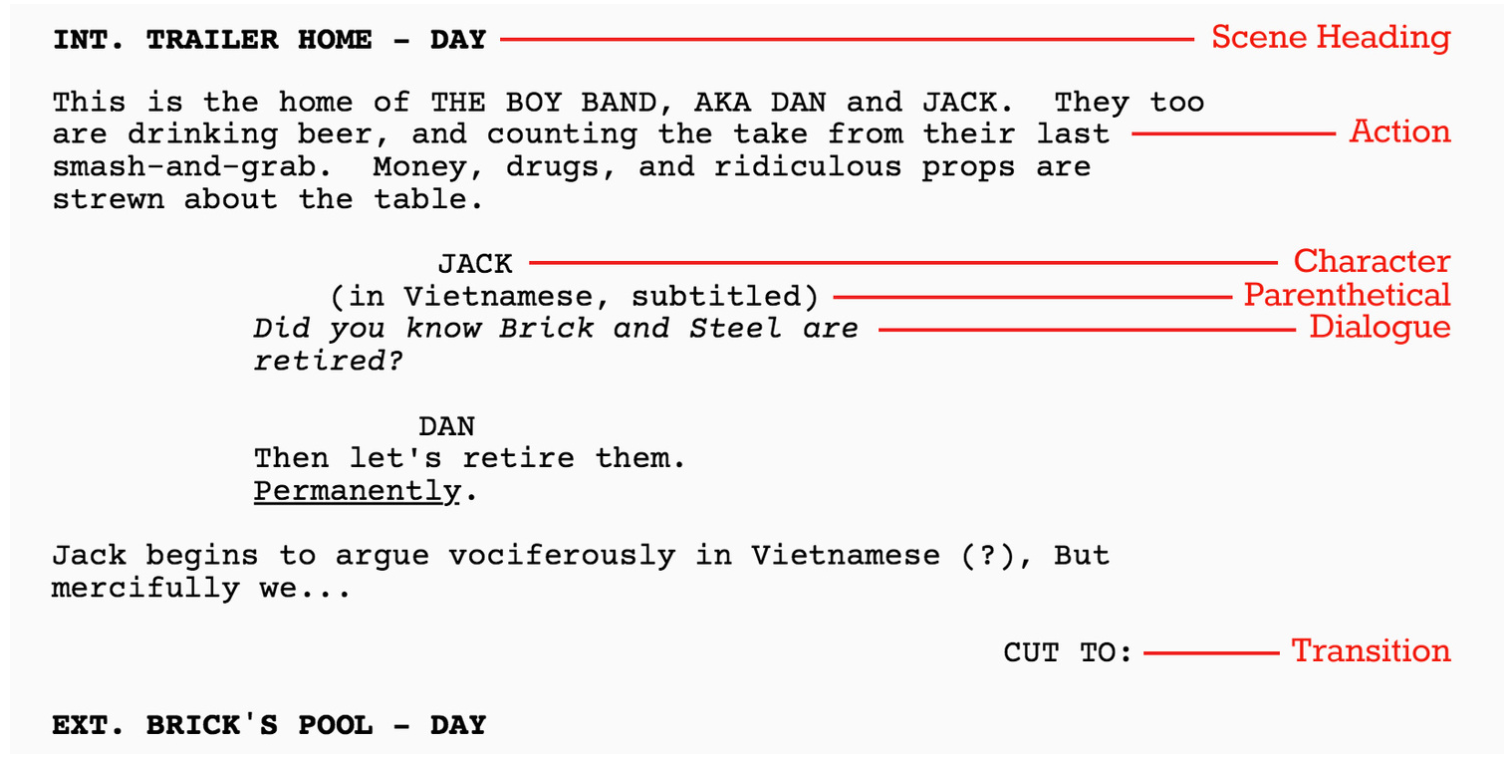}}
   \caption [Caption of MOVSTRUCT]%
   {A screenshot of movie script structure \footnotemark}
    \label{fig:script-structure}
 \end{figure}

The script or screenplay is a written work for an audio-visual medium, \textit{viz.,} film, television series.  It shows the movement, the action, the dialogue of characters and build the storyline along with the plot. 
A standard structure is followed for the script writing to distinguish between characters, action lines, and dialogues. Figure \ref{fig:script-structure} shows a snapshot of a script structure for reference.
 The elements in a screenplay are:
\begin{itemize}[noitemsep] 
    \item \textbf{Scene Heading:} It shows any change in location or time of the scene. It always begins with EXT. (Exterior), INT. (Interior), INT./EXT. (used for scenes inside vehicles). It is always written in uppercase.
\item \textbf{Action:} Description of the scene
\item \textbf{Character:} Name of the character in uppercase. It is followed by the dialogue in
the new line.
\item \textbf{Parenthetical:} A short description placed between character and dialogue to inform
the reader of the character's intent. It is always in parenthesis.
\item \textbf{Dialogue:} To express a character's point-of-view or to understand the interpersonal dynamics between two or more characters. It is always preceded by the character’s name.
\item \textbf{Transition:} Used to indicate a passage from one scene to another. It is also in
uppercase e.g. CUT TO, FADE IN BLACK.
\end{itemize}

\footnotetext{\url{https://slugline.co/basics}}

 \subsection{About Annotation}
 The text scripts for annotation were taken from Internet Movie Script Database (IMSDb)\footnote{\url{https://imsdb.com/}}. The dataset consists of 35 Hollywood movies annotated for bias. The data statistics are given in table \ref{tab:Bias distribution }.
 

\begin{table}[]
\centering
\resizebox{0.5\textwidth}{!}{%
\begin{tabular}{ll}
\hline
\multicolumn{1}{c}{\textbf{Categories}} & \multicolumn{1}{c}{\textbf{Targets}} \\ \hline
Ageism & Kids, Teenagers, Young, Middle aged, Old \\ \hline
Gender & Female, Male \\ \hline
Race/Ethnicity & Black, White, African, English etc. \\ \hline
Religion & Christian, Judaism, Aethist, Islam etc. \\ \hline
LGBTQ & Lesbian, Gay, Bisexual    etc. \\ \hline
Occupation & Defence, Hospitality etc. \\ \hline
\begin{tabular}[c]{@{}l@{}}Other   (Body Shaming,\\  Personality etc.)\end{tabular} & Female, Male \\ \hline
\end{tabular}%
}
\caption{Types of bias categories and a sample of their targets.  }
\label{tab:bias-categories}
\end{table}

\subsubsection{Annotators}
Owing to the complexity of the task, it was decided to work with two specialized annotators with expertise in American history, culture and politics instead of opting for crowdsourcing. Since the annotator's background and life experiences affect the decision making for annotation, special attention was given to minimize the annotator's bias. To make the annotation process more objective, a detailed guideline and a questionnaire set have been curated to assist the annotators (Refer to \hyperref[sec:appendix]{appendix} for a snapshot of guideline and questionnaire). The efficacy of the supporting annotation material was established by asking a new set of annotators to annotate a sample of dataset. 
The Inter Annotator Agreement (IAA) between two expert annotators was computed to establish the impact using kappa score. Table \ref{tab:kappa} shows the cohen kappa score \cite{cohen_kappa} across all labels of annotation. An average kappa score of 0.64 for all possible categories shows a good agreement between the two annotators. However, the agreement for stereotype was low, possibly because of lack of historical knowledge about some communities present in the dataset.
\begin{figure}
  \centering
  \includegraphics[width=\linewidth, height=4.8cm]{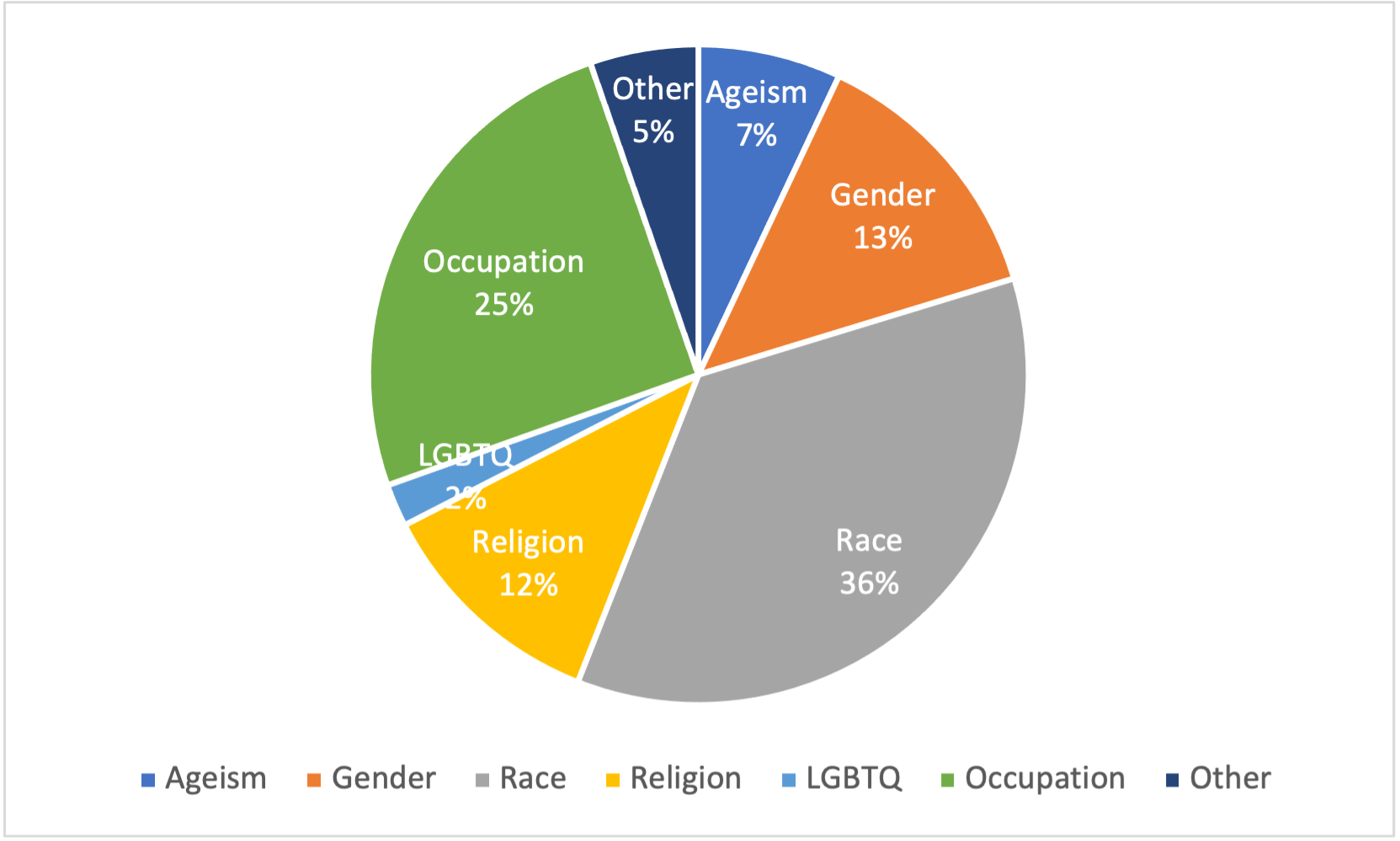}
   \caption{Distribution of social biases across 7 categories. We show percentages of each category annotated in the dataset. }
      \label{fig:bias-category}
 \end{figure}

\subsubsection{Annotation Process} \label{sec:process}

As bias is an inherently complex phenomenon, its analysis, labeling, and subsequent category detection are challenging. The standardised annotation process described here has been modified multiple times in consultation with annotators since the beginning of the task. The granularity of annotation, terminology definition, and bias categories have gone through multiple iterations of revision.


We followed a context-aware layered annotation approach to annotate for multiple tags. The annotation is done at Dialogue Turn (DT) level. Figure \ref{fig:interface} shows a snippet of the interface used for annotation. Each DT contains the following elements:
\begin{itemize}[noitemsep] 
\item \textbf{Scene Description:} Describes the scene for the given dialogue for annotation. 
\item \textbf{Pre-context:} Contains previous 2 dialogues.
\item \textbf{Dialogue:} Current dialogue to be annotated.
\item \textbf{Text to annotate:} Dialogue sentence for annotation.
\item \textbf{Post-context: }Contains following 2 dialogues.
\end{itemize}

For each DT, the labelling was done for \textit{'Text to annotate'}. Each layer labeled is for:

\begin{itemize}[noitemsep] 
\item \textbf{Layer 1:} \textit{Sensitivity.}  Labeled as \textit{Yes or No}.
\item \textbf{Layer 2:} \textit{Stereotype.} Labeled as \textit{Yes or No}. 
\item \textbf{Layer 3:} 7 \textit{category of Bias.} Labeled as \textit{Yes or No}, the target  and reason for bias. A sample of target groups are shown in table \ref{tab:bias-categories}. 
\item 	\textbf{Layer 4:} \textit{Sentiment.}
Labeled as \textit{Positive or Negative}.
\item  \textbf{Layer 5:} \textit{Emotion.}
Labeled for 8 emotion types based on the Plutchik's Wheel of Emotions.   
\item \textbf{Layer 6 :} \textit{Emotion Intensity.} Labeled as \textit{Low, Medium, High}.
\end{itemize}

While annotating, it was found that bias is not always present in an explicit manner. Many times the bias is present based on the identity of the speaker. Thus, the speaker's identity brings in a lot of subtle biases. These constitute implicit biases which are more challenging to identify as it requires identity information about the speaker to be known to annotators. Hence, besides script annotation, metadata about the movie characters are also collected.\\~\\
\fbox{\begin{minipage}{21em}
\small
\textit{\textbf{Action item:}  He giggles at his own joke.}\\
\textit{\textbf{Pre-context:} ITALIAN BOY:  This kid lives around here, but he can't say bread in Italian. CLEMENZA:  That's 'cause he's Jew. Look at those pregnant lips!}\\
\textit{\textbf{Dialogue:} ITALIAN BOY:  Are you a Jewboy?}\\
\textit{\textbf{Text to annotate:} ITALIAN BOY:  Are you a Jewboy?}\\
\textit{\textbf{Post-context:} ITALIAN BOY:  Well, if you're not a Jew, say 'bread' in Italian. See, he can't. CLEMENZA:  Alright, alright, cut it out.}\\

\textit{\textbf{Example 1}:} \textit{\textbf{Bias- }Religion; \textbf{Target-} Jew; \textbf{Reason-} Jewboy is a highly derogatory phrase for boys hailing from Jewish community; \textbf{Sentiment-} Negative; \textbf{Emotion-} Disgust}
\end{minipage}}\\~\\

\textbf{Metadata}: A two-fold metadata is maintained for each movie script to analyse the character graph, and the existence of implicit and explicit biases. 
In \textbf{metadata 1}, we collect the data about the movie such as genre, year of release, director, screenplay writer. Information about the author/director is also collected on the scripts based on book or series.
In \textbf{metadata 2}, we collate information regarding the main characters in the movie such as their age, occupation, gender, religious, racial identity and their role in the movie.

To ascertain the quality of dataset, after annotation, an expert-driven group consisting of 3 members have validated every tag. The inconsistencies in the data were resolved through adjudication. A sample annotation from the dataset is given in Example 1.
\begin{table}[]
\centering
\resizebox{0.3\textwidth}{!}{%
\begin{tabular}{lr}
\hline
\textbf{Labels} & \multicolumn{1}{l}{\textbf{ Kappa Scores}} \\ \hline
Ageism & 0.72 \\ \hline
Gender & 0.54 \\ \hline
Race/Ethnicity & 0.61 \\ \hline
Religion & 0.67 \\ \hline
LGBTQ & 1 \\ \hline
Occupation & 0.47 \\ \hline
Other & 0.49 \\ \hline
\textbf{AVERAGE (all categories)} & \textbf{0.64} \\ \hline
Stereotype & 0.44 \\ \hline
Sensitivity & 0.33 \\ \hline
\textbf{Bias} & 0.71 \\ \hline
\end{tabular}%
}
\caption{Inter-annotator agreement scores using Cohen Kappa for Annotated Labels}
\label{tab:kappa}
\end{table}

\section{Challenges in Annotation}
\label{sec:challenges}

\begin{table*}[]
\resizebox{\textwidth}{!}{%
\begin{tabular}{lll}
\hline
\multicolumn{1}{c}{\textbf{Text to annotate}} & \multicolumn{1}{c}{\textbf{Labels Annotated}} & \multicolumn{1}{c}{\textbf{Reason}} \\ \hline
\begin{tabular}[c]{@{}l@{}}T'CHALLA:  If you were not so stubborn\\  you would make such a great queen.\end{tabular} & \begin{tabular}[c]{@{}l@{}}Bias: Occupation,  \\ Target: Administrative Services\end{tabular} & \begin{tabular}[c]{@{}l@{}}Social perception that great   \\ queens are flexible in their opinion.\end{tabular} \\ \hline
\begin{tabular}[c]{@{}l@{}}NAKIA:  I would make a great queen \\ because I am so stubborn .\end{tabular} & \begin{tabular}[c]{@{}l@{}}Bias: Occupation,  \\ Target: Administrative Services\end{tabular} & \begin{tabular}[c]{@{}l@{}}An individual opinion that great  \\  queens can be rigid in their worldview.\end{tabular} \\ \hline
\begin{tabular}[c]{@{}l@{}}ROTH:  I want my own doctor; fly him in \\ from Miami. I don't trust a doctor who\\  can't speak English\end{tabular} & \begin{tabular}[c]{@{}l@{}}Bias: Race, \\ Target: English\end{tabular} & \begin{tabular}[c]{@{}l@{}}Associating a negative attribute  \\  on the basis of language\end{tabular} \\ \hline
\begin{tabular}[c]{@{}l@{}}MADISON:   Oh, well first off, \\ I have a huge rack.\end{tabular} & \begin{tabular}[c]{@{}l@{}}Bias: Gender, \\ Target: Female\end{tabular} & \begin{tabular}[c]{@{}l@{}}The speaker is objectifying and   \\ equating herself to a sexual commodity.\end{tabular} \\ \hline
\end{tabular}%
}
\caption{Challenging cases from the dataset. Here we have mentioned only the explicit bias examples. Detailed discussions are present in Section \ref{sec:challenges}.}
\label{tab:dt-challenging}
\end{table*}
 
 The annotation process posed some challenges for the annotators. This section discusses the challenges experienced by the annotators. Some of the common challenges faced were:
\begin{itemize}[noitemsep] 
    \item \textit{Implicit Bias} poses a serious challenge to the cause of annotating data as the essence of the underlying bias in the particular utterance is drawn from the entire dialogue turn, character metadata or world knowledge.  Example 2 shows one such case from dataset.\\

\fbox{\begin{minipage}{19em}
\small
\textit{\textbf{Action item:} He addresses that to Neri, who resentfully goes to fetch the Senator a glass of water.\\
\textbf{Pre-context:}MICHAEL:  I trust these men with my life. They are my right arms; I cannot insult them by sending them away. 
SENATOR GEARY: (taking out some medication) Some water.\\
\textbf{Dialogue:}SENATOR GEARY: Alright, Corleone. I'm going to be very frank with you. Maybe more frank than any man in my position has ever spoken to you before.\\
\textbf{Text to annotate:}Maybe more frank than any man in my position has ever spoken to you before.\\
\textbf{Post-context:} SENATOR GEARY: The Corleone family controls two major hotels in Vegas; In another week or so you'll move Klingman out, which leaves you with only one technicality. MICHAEL:  Turnbull is a good man.}\\

\textit{\textbf{Example 2: }\textbf{Bias-} Occupation; \textbf{Target-} Anti-social; \textbf{Reason-} Politicians are assumed to be diplomatic.}\\
\end{minipage}}\\

Here, the dialogue appears as a statement unless we know that the speaker is a \textit{Senator}, a politician by profession that requires diplomatic skill. Hence, this DT is annotated as an Occupation Bias. The bias is coming from world knowledge about a profession.

\item Friends \textit{addressing }each other\textit{ using racial terms} like nigga, black etc. in a friendly way. 

\item There are several instances of \textit{scene level bias }which does not overlap in the dialogues. In such cases the strategy adopted is not to label the bias inherent in the scene as it would not only complicate the process of annotation but also, mislead the model while training process.

\item While tagging \textit{sarcastic comments} (if biased), the sentiment tagged is always negative as incongruity constitutes the key in cases pertaining to sarcasm. 
\item Movie scripts come with inherent biases relating to \textit{gendered roles} and \textit{sexist comments.} In order to standardize the process of labelling self objectification, where the speaker is highlighting himself/herself as an object of sexual desire is to be tagged in positive sentiment. 
\item Sometimes, world knowledge is required to identify bias. Example 3 highlights one such case where the statement is referring to Barack Obama as US President. 
\end{itemize}

\fbox{\begin{minipage}{21em}
\small
\textit{\textbf{Action item:} A large pot is placed atop the gas stove. Everyone eats their gumbo in the kitchen. Laughing talking. Wanda blows out the candles on her cake, cuts it, Hands out everyone a slice. Grandma Bonnie sits in a recliner at the foot of the coffee table.\\
\textbf{Pre-context:} GRANDMA BONNIE: If somebody told me say twenty years ago, that I would live to see it happen, I'd have laughed in their face. OSCAR:  If somebody would have told me that two years ago I would have laughed in their face.\\
\textbf{Dialogue:} DARYL: And here we are. 2008 and its happened.\\
\textbf{Text to annotate:} 2008 and its happened.\\
\textbf{Post-context:} CEEPHUS: But I think it's a sign of the times. The country is in bad enough shape for them to hand it over to a black guy. Look, if Bush hadn't have messed the country up so bad. Would he have even had a shot?}\\

\textit{\textbf{Example 3:}
\textbf{Bias-} Race; \textbf{Target-} Black-Americans; \textbf{Reason-} referring to Obama as president of USA.}\\
\end{minipage}}\\

 Table \ref{tab:dt-challenging} presents some unique challenges faced while labelling. As seen, row 1 and 2 contradict each other but both are taken as Occupation Bias since they are opinionated statements related to the qualities of a Queen. In row 3, the dialogue shows bias related to language. Here,  the bias is both implicit against Spanish and explicit towards the English language. In row 4, the speaker is self objectifying based on gender-specific traits so it is labeled as Gender Bias. These few examples reiterate the fact that bias identification is a complex task in itself.

\section{Experimental Setup}
\label{sec:experiment}
In this section we discuss about different models for detection of social biases and their categories.

\subsection{Data Splits}
As we discussed in previous sections (refer, \ref{sec:process}), our annotation is done at sentence level, i.e., \textit{Text to annotate}. However, due to the low prevalence of bias at the sentence level (table \ref{tab:Bias distribution }), we process the dataset at the dialogue level. A dialogue (speaker: utterance) is considered biased if at least one of the sentences in it has been annotated for any of the 7 bias types. We use the dialogues as input to the model for our various classification tasks. We split the movies based on the count of biased dialogues in them. There were a total of 976 biased dialogues across the span of 35 movies, and the splits are training (488 dialogues; $50\%$), development (213 dialogues; $\sim 22\%$), and test (275 dialogues; $\sim 25\%$). The distribution of different splits across all the categories is mentioned in the table \ref{tab:Bias cat distribution }. We ensured that all the dialogues of the same movie were assigned into the same split to avoid dataset leakage \cite{data-leakage}. By this process, we have 10 movies in the train split and 8, 17 movies respectively in development and test split. 

\begin{table}
\centering
\begin{tabular}{lllll}
\hline
Labels & Sentence Level & Dialogue level \\ \hline
Bias & 1181 (2.40\%) & 976 (3.42\%)  \\
Neutral & 47936 & 27558  \\ 
\hline Total & 49117 & 28534 \\ \hline
\end{tabular}
    \caption{Distribution of Biased sentences and dialogues. }
    \label{tab:Bias distribution }
\end{table}

\begin{table}
\centering
\begin{tabular}{lllll}
\hline
Labels & Train & Dev. & Test \\ \hline
Age Bias & 49 & 22 & 18\\
Gender Bias & 87 & 31 & 56\\
Religion Bias & 62 & 47 & 34 \\
Occupation Bias & 141 & 68 & 105\\
LGBTQ Bias & 11 & 4 & 14\\
Race Bias & 257 & 89 & 97\\
Other Bias & 44 & 9 & 15\\
\hline 
\end{tabular}
    \caption{Distribution of different bias categories across different splits. }
    \label{tab:Bias cat distribution }
\end{table}

\begin{table*}[htp]
\centering
\begin{tabular}{l|lll|lll}
\hline & \multicolumn{3}{c}{ LR } & \multicolumn{3}{c}{ BART-large (SA) }  \\
& $\mathbf{P}$ & $\mathbf{R}$ & $\mathbf{F 1}$ & $\mathbf{P}$ & $\mathbf{R}$ & $\mathbf{F 1}$  \\
\hline 
Race/Ethnicity bias & 0.500 & 0.410 & 0.450 & 0.77 & 0.89 & \textbf{0.83}  \\
Religion bias & 0.226 & 0.259 & 0.241 & 0.86 & 0.67 & \textbf{0.75}  \\
Gender bias & 0.302 & 0.432 & 0.355 & 0.73 & 0.73 & \textbf{0.73}  \\
Occupation bias & 0.321 & 0.464 & 0.380 & 0.59 & 0.48 & \textbf{0.53} \\
Ageism bias & 0.171 & 0.462 & 0.250 & 0.62 & 0.62 & \textbf{0.62} \\
LGBTQ bias & 0.158 & 0.273 & 0.200 & 1.00 & 0.73 & \textbf{0.84} \\
\hline
\end{tabular}
    \caption{Performance of each category on test set. Best F1 scores are in bold. }
    \label{tab:category results }
\end{table*}

\subsection{Methodology}
Due to the shallow presence of biased instances in the data, we use \textit{inductive transfer learning} \cite{ruder-etal-2019-transfer} for our experiments. We use the idea of sequential adaptation, where a model is fine-tuned on a large dataset of related domains before doing the fine-tuning on the target task. Bias detection is done alone in a binary classification framework. As our dataset has been annotated with multiple labels for bias categories, we formulate the category detection task as a multi-label classification problem. For baseline models, we use logistic regression to compare the performances.

\textbf{Logistic Regression (LR):}
We make use of logistic regression with L2 regularization \cite{Nigam99usingmaximum} for both binary classification (bias vs. neutral) and multi-label classification (the bias categories). For multi-label classification, we encode multiple labels per instance at once. We use class weight inversely proportional to the class frequencies for both binary and multi-label training.

\begin{table}
\centering
\begin{tabular}{lllll}
\hline
Models & P & R & F1\\ \hline
LR & 0.53 & 0.71 & 0.51 \\
LR-contrl & 0.52$\pm$0.008 & 0.70$\pm$0.011 & 0.49$\pm$0.021 \\ 
SA & \textbf{0.55} & \textbf{0.81} & \textbf{0.58} \\ 
SA-contrl & 0.55$\pm$0.007 & 0.80$\pm$0.012 & 0.57$\pm$0.017 \\ \hline
\end{tabular}
    \caption{Performance of binary classification[Bias vs. Neutral], on test set, with best numbers in bold. Standard deviations are reported for controlled experiments. }
    \label{tab:binary classification }
\end{table}

\textbf{Sequential Adaptation (SA):}
Because of the heavy skewness of the data, learning directly from it may not give good performances. Hence, we train a model in a sequential adaptive manner \cite{phang2019sentence}. First, we fine-tune the model on a curated dataset of a few related tasks before fine-tuning again on our dataset. We extract \textit{stereotype, unrelated} instances from \cite{nadeem-etal-2021-stereoset} for four categories (gender, race, religion, occupation); group targeted posts from \cite{sap2020social} for race, gender age categories; comment texts with identity\_attack $>$ 0.5 from \cite{jigsaw} where target belongs to one of the categories among race, gender, religion, LGBTQ. We carefully collate $16k$ data instances from these three sources with $6k$ instances for the bias class. 
We use this dataset, as initial training, for binary and multi-label classification. We fine-tune a BART-large \cite{bart} model for both tasks. While fine-tuning later on our dataset, we do a controlled experiment where the number of instances for the neutral class is equal to that of the bias class for binary classification. We report standard deviation across 5 runs for the controlled experiment as we randomly sample the instances for neutral class. We do not include \textit{other class} in the category detection training as there are no publicly available data for all the subcategories (refer table \ref{tab:bias-categories}).

\subsection{Results}
\textbf{Performance Metrics:} 
We report precision (P), Recall (R), F1 score for both binary and multi-label classification in Table \ref{tab:binary classification } and \ref{tab:category results } respectively. Due to the heavy skewness of the data, we rely on the macro values for the classification of bias vs. neutral dialogues.

\textbf{Model Comparison:}
The BART-large (SA) model substantially outperforms logistic regression for bias category detection task (macro F1 score of $0.72$ compared to $0.31$ for LR). However, for bias detection, BART-large shows marginal performance improvement over LR which can be due to the skewed nature of the data. This shows that the category detection task is easier than bias detection.

\subsection{Error Analysis}
We have observed that the category detection model sometimes predicts some extra categories for the dialogue which are not available in the ground truth label. Consider the below dialogue, which has a clear bias towards church (religion), but the model also predicted one extra label as gender, understandably due to the phrase "good woman".\\ 

\fbox{\begin{minipage}{21em}
\small
\textit{\textbf{Sentence :} PHILLIPA: So Stephen, you've been to church with a good woman. Are you feeling holier than thou?\\
 \textbf{Ground Truth :} Religion\\
 \textbf{Prediction :} Gender, Religion}
\end{minipage}}

Occupation bias in the category classification, mostly, has been misclassified as gender bias. The bias classification model, sometimes, assigns the wrong label for neutral dialogues due to the presence of many identity related words/phrases in it. In the below mentioned example, there is no (implicit/explicit) bias towards any identity, but model misclassifies because of the presence of words related to an identity.\\

\fbox{\begin{minipage}{21em}
\small
\textit{\textbf{Sentence :} ORDELL (CONT'D): Oh, that's just my white friend, Louis. He's got nothing to do with my business. We just hangin together. We're on our way to a cocktail lounge.\\
\textbf{Ground Truth :} Neutral \\ \textbf{Prediction :} Bias}
\end{minipage}}

\subsection{Challenges}
Understanding the conversation flow (who say what to whom) in a movie script is a complex phenomenon due to the very high number of dialogue exchanges among multiple characters. Detecting bias at the dialogue level requires the model to understand this flow and also requires understanding the relations among multiple speakers. Sometimes an utterance is biased because of the speaker's demographics. It's quite challenging for state-of-the-art models to capture these nuances correctly. More often, understanding the bias in dialogue requires the other dialogue social context. Providing these external social contexts to AI based models is not a very trivial task. Another major challenge with movie scripts is that the speaker names are not always the named entities (e.g. DEVIL GIRL, GAURD, CLERK). The presence of these identity related words as speaker makes the task more challenging. 

\section{Ethical Consideration}
Our work aims at capturing various social biases in Hollywood movie scripts. We discuss the various challenges we faced in the annotation process, specifically due to the presence of implicit biases in dialogues. The study of social biases comes with ethical concerns of risks in deployment \cite{Ullmann2020}. As these biased dialogues can cause potential harm to any viewer or community, it is required to conduct this kind of research to detect them before release. However, we do not intend to harm the reputation of movie production houses and any other person involved with any of the movies in our dataset. Researchers working on the problem of social bias in fiction, movies, or conversational AI would benefit from the dataset we have created. 

\section{Conclusion and Future Work}
\label{sec:conclusion}
We present a comprehensive dataset of 35 Hollywood movies to identify social biases in movie scripts. The dataset is labeled for \textit{sensitivity, stereotype, identity biases as gender, ageism, race/ethnicity, religion, occupation, LGBTQ, other (body shaming, personality, etc.), the target of bias, sentiment, emotion, emotion intensity, and reason for bias}. The dataset has been benchmarked for bias identification and categorization task using the BART-large model.

Since bias is found to exist only in 3\%  of the dialogues from 35 movies, we have the challenging task of addressing skewness in data. Also, sensitivity labeling was introduced later in the annotation process. We plan to extend sensitivity tags to all the scripts in the dataset. In future, we would like to use the annotated metadata along with dialogue contexts to make the bias detection more robust.


\section{Acknowledgements}

We thank the anonymous reviewers for their insightful comments. Additionally, we thank Sushma Gawande for her contribution in data annotation. We also thank Tamali Banerjee and Sandeep Singamsetty for questionnaire and guideline validation. This research work has been funded by Accenture Labs, India.

\section{References}\label{reference}

\bibliographystyle{lrec2022-bib}
\bibliography{my-references}

\section{Language Resource References}
\bibliographystylelanguageresource{lrec2022-bib}
\bibliographylanguageresource{languageresource}

\appendix
\section{Appendix}
\label{sec:appendix}
\begin{figure*}
  \centering
  \frame{\includegraphics[width=0.85\linewidth, height=5cm]{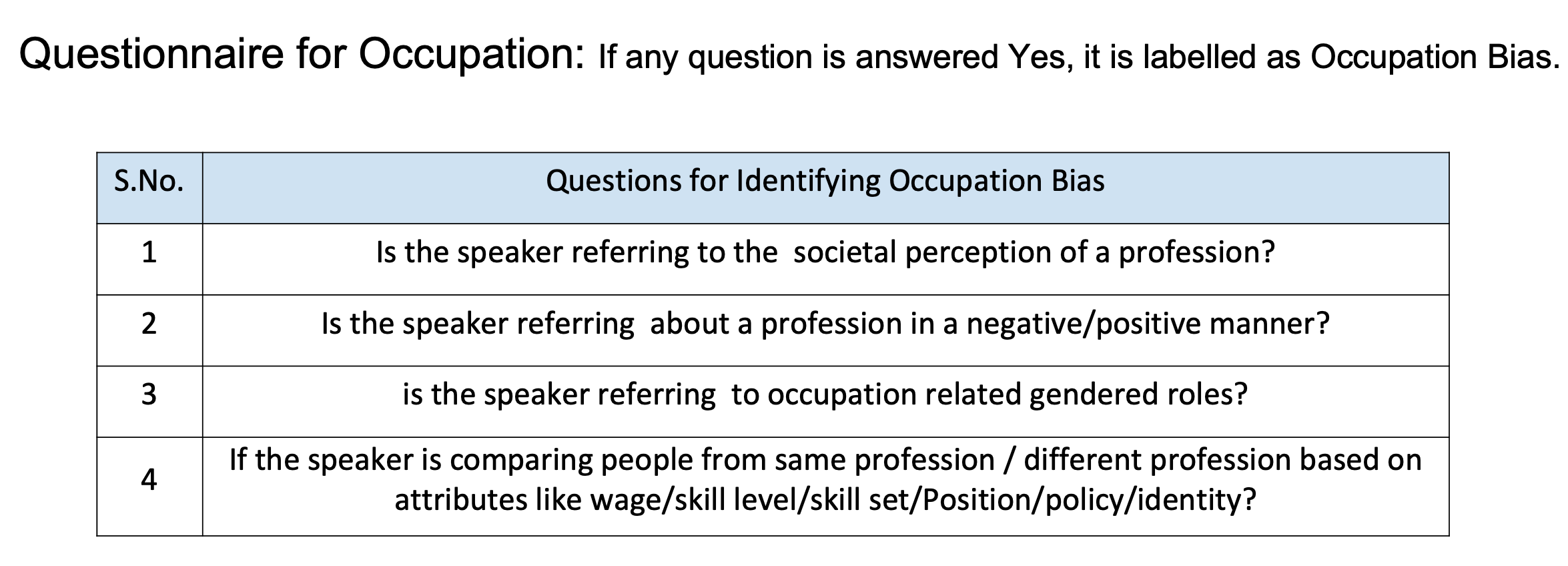}}
   \caption{A snippet of questionnaire to assist the annotators for objective annotation. 
   }
      \label{fig:questionnaire}
 \end{figure*}
 
 \begin{figure*}
  \centering
  \frame{\includegraphics[width=0.7\linewidth, height=6.5cm]{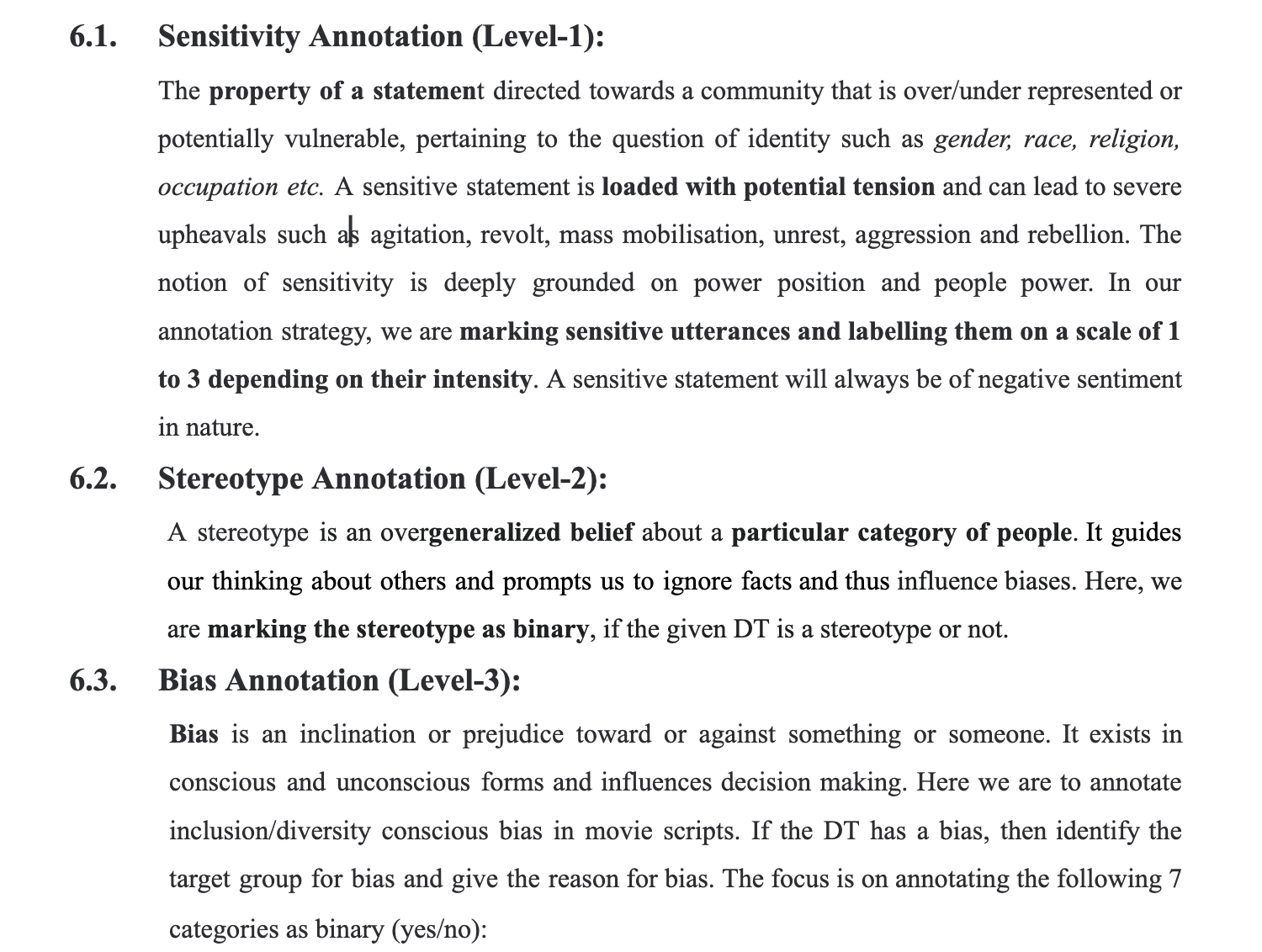}}
   \caption{A snapshot of guidelines used for annotation.
   }
      \label{fig:guide}
 \end{figure*}
 
 \begin{figure*}
  \centering
  \frame{\includegraphics[width=0.6\linewidth, height=4.5cm]{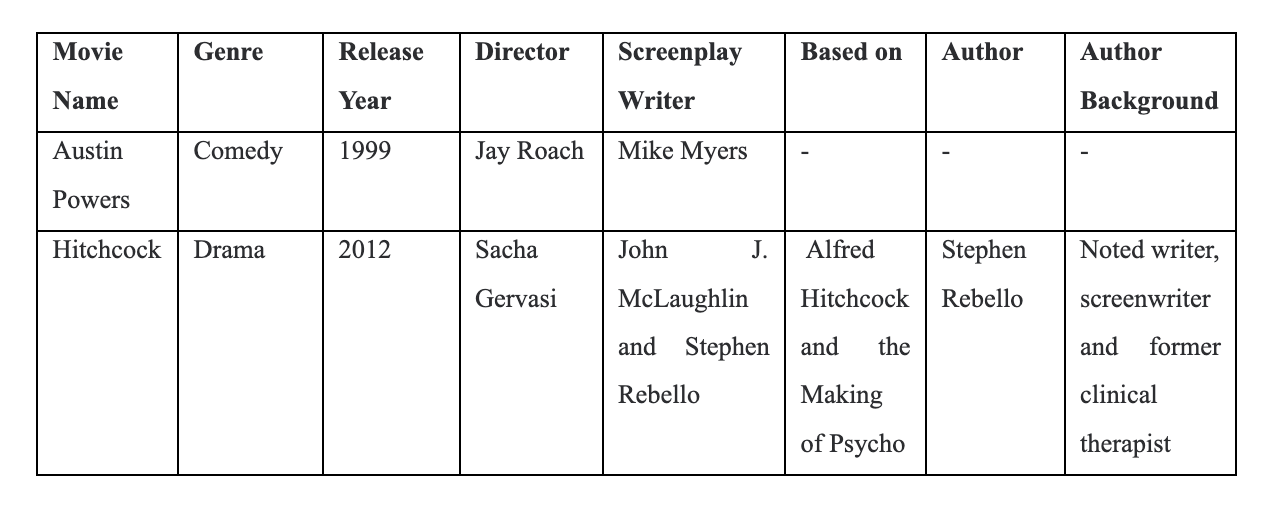}}
   \caption{A snippet of metadata-1. This contains information about the genre, year of release, director, script writer about movie script is collated. 
   }
      \label{fig:meta1}
 \end{figure*}
\subsection{Annotator Details}
Both the annotators were trained and selected through extensive one-on-one discussions. Both of them went through a few days of initial training. They would annotate many examples that would then be validated by our research group and were communicated properly about any incorrect annotations during training. As there are potential adverse side effects of annotating such toxic comments, we used to have regular discussion sessions to ensure they were not excessively exposed to the harmful contents. Both the annotators were middle-aged Asian females and with post-graduation degree in sociology. We took help from two other annotators for validation of guideline and the questionnaire. Both of them have post-graduation degree in computer science.

\subsection{Training Details}
We finetune BART-large models with batch size of 32 for both tasks. For sequential adaptation, we finetune the models for five epochs on the curated dataset before re-finetuning them for two epochs on our dataset. Max token length of 128 is used for both the classifications. We also use a dropout layer in our model. We experiment with learning rates of $\{2e-5,3e-5, 4e-5, 5e-5\}$. AdamW \cite{adamw} optimizer with epsilon = 1e-08, clipnorm = 1.0, 200 warm up steps are used. Experiments were run with a single GeForce RTX 2080 Ti GPU.

All of our implementations uses Huggingface’s
transformer library \cite{wolf2020huggingfaces}.

\subsection{Snapshot of Guideline}
\begin{figure*}
  \centering
  \frame{\includegraphics[width=0.6\linewidth, height=6cm]{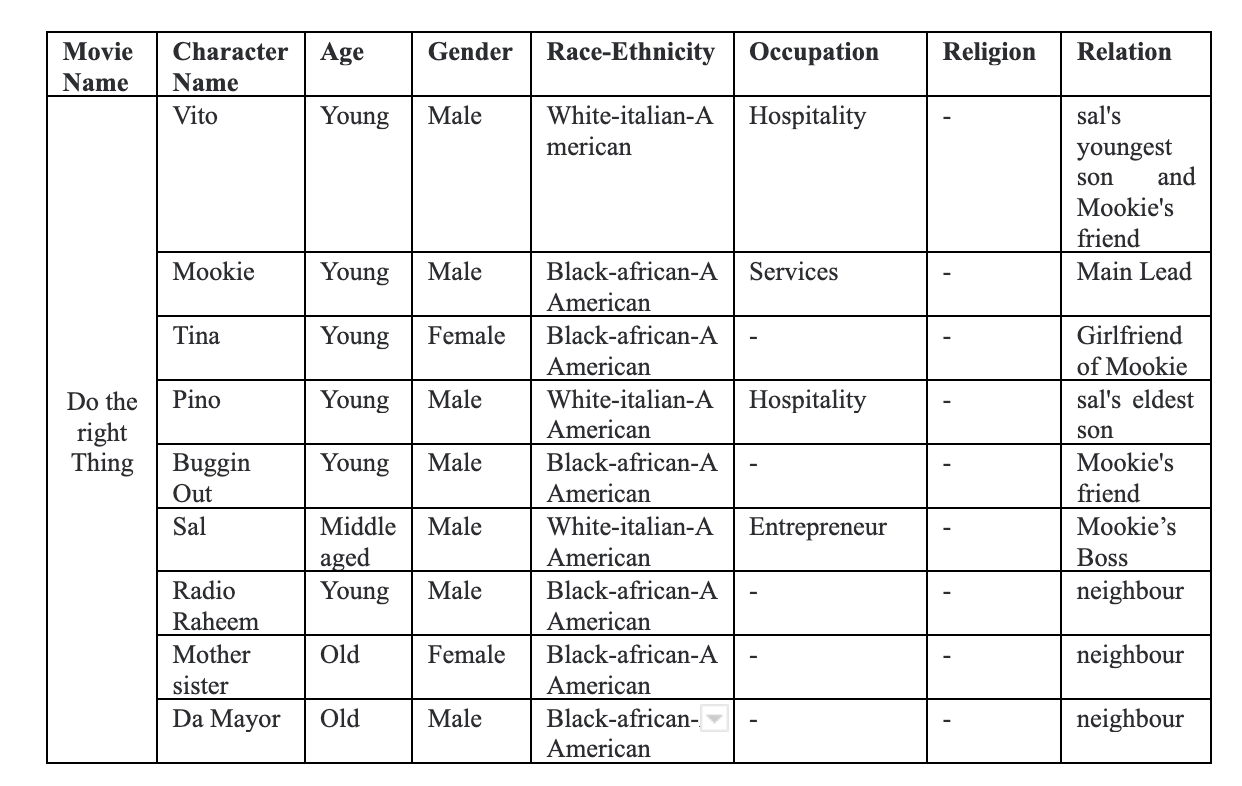}}
   \caption{A snippet of metadata-2. This has information related to movie character is collated, \textit{viz.,}character's gender, age, race, occupation etc.
   }
      \label{fig:meta2}
 \end{figure*}

Bias being a subjective phenomena, a questionnaire and a detailed annotation guideline was created to assist the annotators for labelling the biases. These supporting guidelines have evolved based on continous feedback from the annotators. Figure \ref{fig:questionnaire} shows a snapshot of the questionnaire used to check the presence of bias. Figure \ref{fig:guide} is a snapshot from the guideline used for annotation.  Figure \ref{fig:meta1} and \ref{fig:meta2} shows the metadata 1 and 2 structure respectively. These metadata is used to understand implicit biases which can come from characters background. Table \ref{tab:occupation-targets} shows the list of occupation targets and the list of occupations included under each target. The complete questionnaire and guideline can be accessed at \url{https://github.com/sahoonihar/HIBD_LREC_2022}

\begin{table*}[]
\resizebox{\textwidth}{!}{%
\begin{tabular}{lll}
\hline
\multicolumn{1}{c}{\textbf{S.No.}} & \multicolumn{1}{c}{\textbf{Occupation Target}} & \multicolumn{1}{c}{\textbf{Occupations Included}} \\ \hline
1 & Healthcare & doctor, nurse,   counsellor, etc. \\ \hline
2 & Hospitality & hotel manager, chef,   bartender, cook, wedding planner etc. \\ \hline
3 & Anti-social & Gangsters, thieves,   con-artists, drug dealers, etc. \\ \hline
4 & Defence & Military officers,   Intelligence (CIA, Mossad, RAW, ISI), National Security, etc. \\ \hline
5 & Entertainment & actors, dancers,   musician, painter, clowns, etc. \\ \hline
6 & Politicians & government office   bearers \\ \hline
7 & Administrative services & Kings, queens, royal blood, presidents, prime-minister,   minister, senator, etc. \\ \hline
8 & Financial and management services & bankers, traders,   managers, etc. \\ \hline
9 & Entrepreneurs & self-employed,   businessman, landlords, etc. \\ \hline
10 & Security & police, watchman,   guards, bounty hunters, etc. \\ \hline
11 & Academia and Research & scientist, professor,   teacher, etc. \\ \hline
12 & Sports & players, sports   manager, referees, coaches, gladiator, etc. \\ \hline
13 & Services & delivery boy, driver,   labourer, desk jobs/clerks, prostitutes, negotiater, secretary, etc. \\ \hline
14 & Technical Services & Engineering services,   mechanics, etc. \\ \hline
15 & Fitness/Sports Instructor & Yoga Trainer, Gym   Trainer, referees, umpires, etc. \\ \hline
16 & Journalist/Press & Media, News Reporters,   Columnist, etc. \\ \hline
17 & Legal Aid & Lawyers, District   Attorney (DA), Judges, Chief Justice, Justice, etc. \\ \hline
18 & Religious Services & Pope, Priests, Nun,   Father, etc. \\ \hline
\end{tabular}%
}
\caption{List of Occupation Bias targets and the occupations included under it.}
\label{tab:occupation-targets}
\end{table*}

\end{document}